\documentclass[letterpaper, 10 pt, conference]{ieeeconf}
\usepackage{times}
\usepackage{graphicx}

\usepackage{cite}

\usepackage{multicol}
\usepackage[bookmarks=true]{hyperref}
\usepackage[normalem]{ulem}
\usepackage{xspace}
\usepackage{subcaption}
\usepackage{booktabs}
\usepackage{siunitx}
\usepackage{etoolbox}
\AtBeginEnvironment{abstract}{\sisetup{mode=text, reset-text-series=false}}
\usepackage{balance} 
\usepackage{eso-pic}
\hypersetup{hidelinks=true}

\usepackage{todonotes}
\usepackage{graphicx}
\usepackage{xcolor}
\usepackage{float}
\usepackage{amsmath,amssymb}
\usepackage[acronym]{glossaries}
\glsdisablehyper

\newacronym{mppi}{MPPI}{Model Predictive Path Integral}
\newacronym{esdf}{ESDF}{Euclidean Signed Distance Field}
\newacronym{vio}{VIO}{Visual-Inertial Odometry}
\newacronym{slam}{SLAM}{Simultaneous Localization and Mapping}
\newacronym{mav}{MAV}{Micro Air Vehicle}
\newacronym{mpc}{MPC}{Model Predictive Controller}
\newacronym{gpu}{GPU}{Graphics Processing Unit}
\newacronym{phtree}{PH-Tree}{PATRICIA-hypercube-tree}
\newacronym{qsm}{QSM}{quantitative structure model}
\newacronym{sqp}{SQP}{sequential quadratic programming}

\begin{document}
\makeatletter
\providecommand{\bstctlcite}[1]{%
  \@bsphack
  \@for\@citeb:=#1\do{%
    \edef\@citeb{\expandafter\@firstofone\@citeb}%
    \if@filesw\immediate\write\@auxout{\string\citation{\@citeb}}\fi}%
  \@esphack}
\makeatother
\bstctlcite{IEEEexample:BSTcontrol}

\newcommand{\secref}[1]{Sec.\,\mbox{\ref{#1}}}
\newcommand{\figref}[2][]{Fig.\,\hyperref[#2]{\ref*{#2}#1}}
\newcommand{\tabref}[1]{Tab.\,\ref{#1}}

\newcommand{\simmode}[1]{lockstep-\textsc{#1}\xspace}

\newcommand{\Mfreespace}{$\mathcal{F}$\xspace}
\newcommand{\Moccupiedspace}{$\mathcal{O}$\xspace}   
\newcommand{\Munknownspace}{$\mathcal{U}$\xspace}
\newcommand{\freespace}{\mathcal{F}}
\newcommand{\occupiedspace}{\mathcal{O}}
\newcommand{\unknownspace}{\mathcal{U}}

\newsavebox{\panelbox}
\newcommand{\figpanel}[4][1.0]{%
  \sbox{\panelbox}{\includegraphics[scale=#1]{#2}}%
  \begin{minipage}[t]{\wd\panelbox}%
    \centering
    \usebox{\panelbox}%
    \captionsetup{margin=5pt}%
    \caption{#3}%
    \label{#4}%
  \end{minipage}%
}

\title{\LARGE \bf
Low-Latency Mapless ESDF Queries for \\ MAV Reactive Collision Avoidance}

\author{Leonard Frei{\ss}muth, Cedric Le Gentil, Helen Oleynikova, and Stefan Leutenegger}
\AddToShipoutPictureFG*{%
  \AtPageUpperLeft{\put(\LenToUnit{54pt},\LenToUnit{-30pt}){\parbox[t]{504pt}{\centering\footnotesize
    This work has been submitted to the IEEE for possible publication. Copyright may be transferred without notice, after which this version may no longer be accessible.}}}%
  \AtPageLowerLeft{\put(\LenToUnit{54pt},\LenToUnit{30pt}){\parbox[b]{245pt}{\footnotesize
    All authors are with the Mobile Robotics Lab, ETH Zurich.\\ \{lfreissmuth, clegentil, oelena, sleutenegger\}@ethz.ch}}}}

\maketitle
\thispagestyle{empty}
\pagestyle{empty}

\begin{abstract}
    Reactive collision avoidance in unknown, cluttered environments must perceive obstacles and correct motion within milliseconds.
    Existing methods rely on maps for \emph{high-rate collision queries}, but their acceleration structures require pre-processing.
    That cost is amortized in long-horizon planning, where a map serves millions of queries; reactive planning, however, is a near-field, short-horizon problem with potentially only hundreds of queries per measurement, where building a map may not pay off.
    We propose a mapless, \gls{gpu}-accelerated \gls{esdf} query with negligible pre-processing that operates directly on posed depth images, with resolution set by pooling the image rather than by a voxel size.
    On an embedded \gls{gpu} it is simultaneously more accurate and lower in total latency than every voxel-grid configuration we measured between \SIlist{2;20}{\centi\meter}, at one to two orders of magnitude less memory at matched resolution.
    With our query integrated into a \gls{mpc}, simulated runs in laser-scanned forests at up to \SI{9}{\meter\per\second} collide in \SI{0.4}{\percent} through dense branches, against \SI{5.9}{\percent} with the best voxel grid and \SI{5.5}{\percent} with the best mapless baseline at our resolution, whose failures we attribute to voxel resolution and total latency respectively.
    We deploy the system onboard at two forest sites and fly at commanded speeds of up to \SI{4}{\meter\per\second}.
\end{abstract}
\glsresetall

\section{Introduction}

For fast \gls{mav} flight in cluttered environments, reactive collision avoidance may require obstacle perception and motion correction within milliseconds.
The local planner must therefore rely on the latest sensor data for fine, near-field obstacles such as branches emerging from the occlusion of nearer geometry.
Standard approaches integrate sensor measurements into a map---such as occupancy grids \cite{hornung_octomap_2013,funk_multi-resolution_2021} or distance fields \cite{oleynikova_voxblox_2017,millane_nvblox_2024}---and query that map for collision.
For a reactive safety layer, the relevant quantity is the \textit{total latency} from a new depth measurement to a collision answer.
It comprises two terms: the \textit{pre-processing} required to build or update the map, including any acceleration structure over it, and the \textit{query time} itself, which grows with the number of queries.
A map trades constant pre-processing cost for lower per-query cost, which pays off in long-horizon planning where a measurement may be reused for millions of queries.
Reactive planning, however, is a near-field, short-horizon problem: the controller queries only its immediate neighborhood, on the order of $10^2$ times per sensor update.
In this regime the pre-processing term, rather than the per-query term, dominates total latency, which raises the question whether a map pays off at all.

\begin{figure}[t]
    \centering
    \includegraphics[width=\linewidth, trim=0 0 0 0, clip]{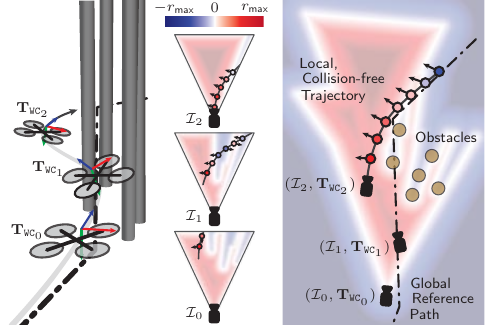}
    \caption{We represent the environment as a set of posed depth images $\mathcal{I}_t$ and compute distances and gradients for a query in every image in parallel before merging them. The dense field in the background is illustrative only, since we evaluate the field at query points alone, and all algorithms work in 3D although we draw in 2D.}
    \label{fig:main-motivation}
\end{figure}

Voxel maps are a common choice for such local planning, but their fixed voxel size has to be traded off twice.
A depth camera resolves finely nearby and coarsely far away, so a fixed voxel size under-represents the near field or over-represents the far field, and finer voxels cost more compute and memory.
Prior mapless methods avoid these tradeoffs by retaining posed depth images, querying each frame through a per-frame KD-tree \cite{florence_nanomap_2018,ji_mapless-planner_2020} or an occlusion-aware pyramid \cite{bucki_rectangular_2020}.
We instead evaluate depth pixels directly in parallel on the \gls{gpu}, with local depth minima and maxima for pruning as the only pre-processing.
The resulting sensor-space \gls{esdf} (\figref{fig:main-motivation}) keeps the sensor's range-dependent resolution, controlled by pooling, returns distances and gradients, handles unknown space optimistically or conservatively, and carries a covariance-based pose-uncertainty margin.
On a Jetson Orin NX it answers the $400$ queries of our reactive planner in \SI{6.2}{\milli\second} at full resolution, one to two orders of magnitude below KD-tree and mesh baselines, and at that setting both more accurate and lower in total latency than every voxel grid we measured, at the least memory at matched resolution (\secref{sec:exp-runtime}).
In closed-loop simulation through laser-scanned forests at up to \SI{9}{\meter\per\second}, runs of a \gls{mpc} are safer with our query than with voxel grids, which fail through resolution, or with KD-trees or meshes, which fail through total latency (\secref{sec:exp-sim}).
We deploy the system in ten forest flights at commanded speeds of up to \SI{4}{\meter\per\second}.

Our contributions are:
\begin{itemize}\csname @beginparpenalty\endcsname=10000 %
    \item A parallel sensor-space distance query algorithm that fuses posed depth views, with pooling-controlled resolution and conservative or optimistic handling of unknown space, benchmarked on an embedded \gls{gpu} for total latency, accuracy, and memory.
    \item A covariance-based uncertainty margin that inflates obstacles by the odometry uncertainty.
    \item A closed-loop evaluation comparing voxel-grid, KD-tree, mesh, and our representation in laser-scanned forests behind one \gls{mpc} on the embedded platform, separating the effects of total latency and resolution, and a demonstration of the running system in ten real-world forest flights at commanded speeds of up to \SI{4}{\meter\per\second}.
\end{itemize}

\section{Related Work}

\subsection{Reactive Collision Avoidance}
We categorize reactive methods as corridor-based, predictive, or learned.
We assume static geometry over the buffered horizon, as prior mapless methods do \cite{florence_nanomap_2018,ji_mapless-planner_2020}; moving obstacles require motion prediction \cite{lindqvist_nonlinear_2020} and are out of scope.

\subsubsection{Corridor-Based Methods}
\label{sec:corridor-methods}
These optimize a trajectory \cite{preiss_trajectory_2017} within overlapping convex sub-volumes of free space, polyhedra \cite{ren_safety-assured_2025,tordesillas_faster_2021} or spheres \cite{ren_bubble_2022,ji_mapless-planner_2020}, tracked by lower-level controllers, and reach \SI{20}{\meter\per\second} in sparse forests with LiDAR \cite{ren_safety-assured_2025}.
Without a prior map they can be trapped in local minima \cite{meijer_pushing_2024}.
Our query targets the reactive layer beneath a global planner and can alternatively serve these optimizers, which query the neighborhood sparingly.

\subsubsection{Predictive Methods}
\gls{mppi} scores perturbed rollouts on the \gls{gpu} \cite{minarik_model_2024,zhai_pa-mppi_2025} but needs large sample counts, making collision queries the bottleneck; \glspl{mpc} \cite{nguyen_model_2021} exploit gradients to converge with fewer queries.
Their collision constraints use a local map \cite{zhang_perception-aware_2018}, a learned latent representation \cite{jacquet_n-mpc_2024,jacquet_neural_2025}, or (neural) control barrier functions \cite{harms_neural_2024,harms_safe_2025}.
EGO-Planner \cite{zhou_ego-planner_2021} reports millisecond optimization times, yet on a separately maintained map whose update bounds its total latency, as it does for Fast-Planner \cite{zhou_robust_2019} and chance-constrained planning over voxel grids \cite{harithas_cco-voxel_2022}.
All of them could consume our query, and we adopt an \gls{mpc} to demonstrate it.

\subsubsection{Learning-Based Methods}
Learned policies map depth to commands, distilled from a privileged planner \cite{loquercio_learning_2021}, learned from demonstrations \cite{kim_rapid_2025}, or trained through differentiable physics \cite{zhang_back_2024}.
They avoid the map at fixed inference latency, at the price of training data, sim-to-real transfer, and a policy tied to its domain and vehicle that cannot be constrained geometrically, whereas our query is training-free and returns distances and gradients usable as hard constraints, as a safety filter beneath a policy, or as supervision.

\begin{figure}[t]
    \centering
    \includegraphics[width=\linewidth]{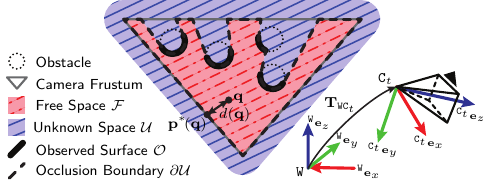}
    \caption{The robot's surroundings divide into known free space \Mfreespace, the observed surface \Moccupiedspace, and unknown space \Munknownspace, and our representation lives in the image space of camera frame $\mathtt{C}_t$.}
    \label{fig:problem-formulation}
\end{figure}

\subsection{Local Map Representations}

\subsubsection{Point Cloud Integrations}
Point clouds dominate for LiDAR sensors and are queried through KD-trees \cite{muja_scalable_2014,bentley_multidimensional_1975}, made incremental \cite{cai_ikd-tree_2021} or rebalancing-free \cite{zaschke_ph-tree_2014}, or interpolated into continuous fields \cite{le_gentil_accurate_2024}.
Maintaining the structure is a constant overhead that dominates total latency at low query counts, and unknown space must be recovered separately \cite{liu_planning_2017,wang_fast_2025,akin_computing_2015}, whereas our query represents it natively.

\subsubsection{Voxel Grids}
Voxel grids integrate measurements into occupancy or distance fields and model occlusion explicitly \cite{oleynikova_signed_2016}.
Octrees \cite{hornung_octomap_2013,vespa_efficient_2018,funk_multi-resolution_2021} curb the cubic memory growth at the cost of traversal overhead.
Voxel hashing and incremental \gls{esdf} updates make fixed-resolution integration efficient on the CPU \cite{duberg_ufomap_2020,oleynikova_voxblox_2017,han_fiesta_2019} and, with NVBlox \cite{millane_nvblox_2024}, on the \gls{gpu}.
Yet cells remain fixed-size, which is memory-inefficient at high resolution, and the per-frame integration can dominate total latency at low query counts.

\subsubsection{Mapless Methods}
Mapless methods query in sensor space.
RAPPIDS \cite{bucki_rectangular_2020} partitions the free space of the most recent image into pyramids for a binary collision test that provides neither distance nor gradient, and a single image gives a limited field of view \cite{lopez_aggressive_2017}.
A history of posed images, queried per frame through KD-trees \cite{florence_nanomap_2018,ji_mapless-planner_2020}, covers more.
We keep this buffer but replace the per-frame structures with a direct pixel-space query at negligible pre-processing cost that, unlike prior mapless methods, returns a \emph{signed} distance with gradients under either handling of unknown space.

\begin{figure*}[t]
    \centering
    \includegraphics[width=0.23\textwidth]{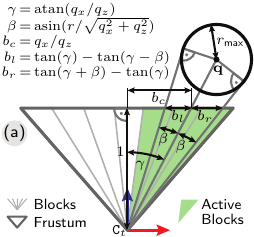}\hfill%
    \includegraphics[width=0.23\textwidth]{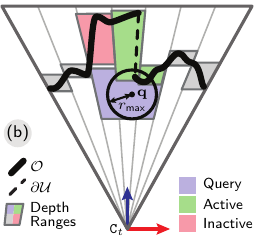}\hfill%
    \includegraphics[width=0.23\textwidth]{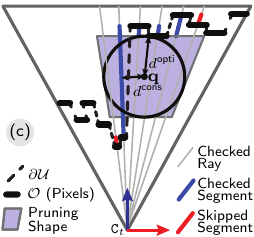}\hfill%
    \includegraphics[width=0.23\textwidth]{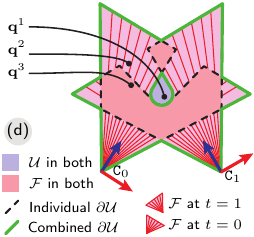}
    \caption{Single-frame query and merging. (a)~The truncation sphere of radius $r_\text{max}$ around $\mathbf{q}$ projects to a box of pixels, and only the blocks it covers are searched. (b)~Of these, a block is active when its stored depth range intersects $[\,q_z-r_\text{max},\,q_z+r_\text{max}\,]$, and all other blocks are skipped. (c)~Occlusion boundaries as ray segments from a pixel's depth to that of its neighbor nearest to $\mathbf{q}$, searched within $d^{\,\text{opti}}(\mathbf{q})$ for $d^{\,\text{cons}}(\mathbf{q})$. (d)~Max-pooling of $\phi_t(\mathbf{q})$ across views in the three cases of \secref{sec:merging}, with only \Munknownspace views at $\mathbf{q}^1$, mixed views at $\mathbf{q}^2$, and only \Mfreespace views at $\mathbf{q}^3$.}
    \label{fig:truncation-overview}
\end{figure*}

\section{Method}

\subsection{Problem Formulation}

We seek an \gls{esdf} function $\phi(\mathbf{q})$ whose absolute value is the distance $d(\mathbf{q})$ from a query point $\mathbf{q} \in \mathbb{R}^3$ to the boundary $\partial\mathcal{F}$ of free space.
The sign of $\phi$ encodes whether $\mathbf{q}$ lies in free space \Mfreespace or unknown space \Munknownspace (\figref{fig:problem-formulation}).
We also seek the point $\mathbf{p}^*(\mathbf{q})$ closest to $\mathbf{q}$ on the boundary of \Mfreespace, from which the gradient follows whenever $\mathbf{p}^*(\mathbf{q})$ is unique:
\begin{equation}
    \nabla \phi(\mathbf{q}) = \operatorname{sgn}(\phi(\mathbf{q}))\, \frac{\mathbf{q} - \mathbf{p}^*(\mathbf{q})}{\lVert\mathbf{q} - \mathbf{p}^*(\mathbf{q})\rVert_2}.
    \label{eq:gradient-from-argmin}
\end{equation}
For low total latency the method should require minimal pre-processing and exploit \gls{esdf} truncation to prune the search.

\subsection{Representation}
\label{sec:representation}

We represent the environment with an $N$-slot circular buffer of posed depth images $\mathcal{I}_t$, each storing depth measurements $z_t(u,v)$ at pixels $u,v$ together with a pose $\mathbf{T}_{\mathtt{W}\mathtt{C}_t}$ that transforms its camera frame $\mathtt{C}_t$ at timestamp $t$ into the world frame $\mathtt{W}$.
We partition each image into \emph{blocks} of $32\times16$ pixels, motivated by \glspl{gpu}' hardware accelerators.
For broad coverage despite a high frame rate, one slot always holds the latest frame while the other $N-1$ hold keyframes, of which a new one replaces the oldest once the pose differs sufficiently in translation, rotation, or time, and all slots are then treated identically and indexed by timestamp $t$.
To query a distance, we evaluate each frame individually (\secref{sec:single-frame}), optionally handle occlusion boundaries $\partial \mathcal{U}$, merge the per-frame results (\secref{sec:merging}), and account for uncertainty in the pose estimate (\secref{sec:uncertainty}).
Here $\partial\mathcal{U}$, the boundary of unknown space, comprises both depth discontinuities that occlude the volume behind them and the limits of the camera frustum, beyond which space is equally unobserved.

The resolution is the pixel footprint $\frac{z}{f_x}$ at depth $z$ and focal length $f_x$, which we coarsen by min-pooling the image with a $p\times p$ kernel on insertion.
Keeping the nearest depth in the kernel inflates fine details rather than losing them, so pooling trades latency for accuracy conservatively.

\begin{figure*}[t]
    \centering
    \includegraphics[width=\textwidth]{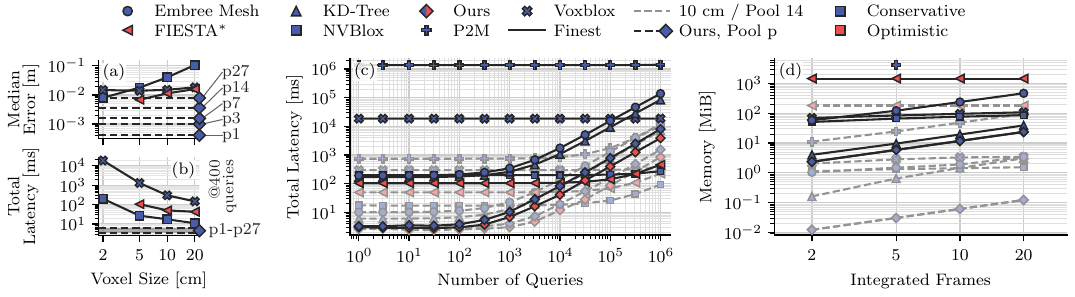}
    \caption{Benchmarks on the Orin NX with $N=1$ frame in (a) and $N=5$ in (b) and (c). (a)~Median absolute distance error against a full-resolution reference, over voxel size for the grids and as dashed lines for our pooling levels $p$, whose coarse settings match the finest grid. (b)~Total latency of $400$ queries from the start of the last integration to the last answer, where every pooling level of ours lies below the fastest grid. (c)~Total latency over the number of queries at each method's finest setting (solid) and at \SI{10}{\centi\meter} or the matching pooling level (dashed). (d)~Map memory over the integrated frames. $^\ast$FIESTA runs at \SI{5}{\centi\meter} at its finest setting, all other voxel grids at \SI{2}{\centi\meter}.}
    \label{fig:res-frontier}
\end{figure*}

\subsection{Single-Frame Query}
\label{sec:single-frame}

To query the \gls{esdf} $\phi$, we project the query point $\mathbf{q}$ into each camera frame $\mathtt{C}_t$ and evaluate all posed depth images in parallel.
For simplicity, we drop the timestamp in this section and write the query point as $\mathbf{q} = \begin{bmatrix} q_x & q_y & q_z \end{bmatrix}^\top$.
Intuitively, the distance we seek is the radius of the largest sphere around $\mathbf{q}$ that contains neither an observed surface \Moccupiedspace nor an occlusion boundary $\partial\unknownspace$, treating unobserved space as occupied.
We obtain the radius in two steps, first for the surface and then for the occlusion boundary, by finding the nearest point back-projected from a depth pixel and clipping at the truncation radius $r_\text{max}$.
For efficiency we restrict the search to \emph{active} blocks that may contain such a point (\secref{sec:truncation-mechanism}) and evaluate their pixels in parallel.

\subsubsection{Truncation Mechanism}
\label{sec:truncation-mechanism}
To prune the search space, we record the minimum and maximum depth within each block whenever a new image arrives, which, besides memory management, is the only source of pre-processing latency.
These per-block depth ranges let us prune most pixels at query time using a sphere centered on $\mathbf{q}$ with radius $r_\text{max}$:
we mark a block \textit{active} only if (i)~one of its pixels lies in the bounding box of the query sphere's projection (\figref[a]{fig:truncation-overview}) and (ii)~its stored depth range intersects $[\,q_z-r_\text{max},\,q_z+r_\text{max}\,]$ (\figref[b]{fig:truncation-overview}).
Every other block lies farther than $r_\text{max}$ from $\mathbf{q}$.

\subsubsection{Nearest Observed Surface}
Within the active blocks we search for the back-projected depth pixel nearest to $\mathbf{q}$, which gives the distance to the nearest observed surface.
We encode each candidate pixel in a single 64-bit key whose most significant 32 bits hold the squared distance from $\mathbf{q}$ to the back-projected 3D point as an IEEE-754 float and whose lower bits carry the coordinates $u$ and $v$.
Because non-negative IEEE-754 floats keep their ordering when reinterpreted as integers, the nearest pixel over the whole image follows from a single hardware-accelerated \texttt{atomicMin} over these keys.
Back-projecting the winning pixel recovers the closest point $\mathbf{p}^{\,\text{opti}}(\mathbf{q})$ at no extra cost, and clipping its distance to $r_\text{max}$ gives $d^{\,\text{opti}}(\mathbf{q})$.

\subsubsection{Occlusion Boundaries}
The second step accounts for the boundary $\partial \unknownspace$ of unknown space, treating $\unknownspace$ as occupied, by modeling occlusion boundaries as segments on the camera rays associated with depth pixels.
For a query point $\mathbf{q}$, each segment is bounded by the pixel's measured depth and the depth of the adjacent pixel whose image location is closest to the projection of $\mathbf{q}$.
We again find active blocks, now with truncation radius $d^{\,\text{opti}}(\mathbf{q})$, take for each segment in them the point closest to $\mathbf{q}$ (\figref[c]{fig:truncation-overview}), and reduce these with the same 64-bit keys, whose lower bits now identify the winning segment, to $d^{\,\text{cons}}(\mathbf{q})$ and $\mathbf{p}^{\,\text{cons}}(\mathbf{q})$.

Reporting $d^{\,\text{opti}}$ directly and skipping this step yields a purely \emph{optimistic} query that treats unknown space as free and saves nearly half the query time, while the full $d^{\,\text{cons}}$ is \emph{conservative}, so one representation serves planners of either kind or, as in \cite{tordesillas_faster_2021, ren_safety-assured_2025}, both at once.

With the timestamp restored, each frame $t$ thus yields a signed distance $\phi_t(\mathbf{q}) = \operatorname{sgn}\!\big(z_t(u,v) - q_z\big)\, d_t(\mathbf{q})$, where the sign, obtained from depth $z_t(u,v)$ at the pixel where $\mathbf{q}$ projects, is positive in \Mfreespace and negative in \Munknownspace.

\subsection{Merging Multiple Views}
\label{sec:merging}

To merge the per-frame signed distances $\phi_t(\mathbf{q})$, we take their maximum, so a free-space observation of $\mathbf{q}$ overrides an unknown-space observation from a frame in which $\mathbf{q}$ is occluded.
As illustrated in \figref[d]{fig:truncation-overview}, this yields three cases:
(i) if both values are negative, both frames report \Munknownspace and the larger lies closer to $\partial\unknownspace$;
(ii) if they differ in sign, the frame that observed $\mathbf{q}$ in \Mfreespace should dominate;
(iii) if both are positive, the maximum selects the larger clearance, which also clears the positive side of occlusion boundaries.

The merged field is therefore obtained by max-pooling over the signed distances with the closest point of the winning timestamp, $t^\star = \operatorname*{arg\,max}_{t}\, \phi_t(\mathbf{q})$, $\phi(\mathbf{q}) = \phi_{t^\star}(\mathbf{q})$, and $\mathbf{p}^*(\mathbf{q}) = \mathbf{p}_{t^\star}(\mathbf{q})$, so $\mathbf{q}$ is in \Mfreespace if it is free in \textit{any} frame.

\subsection{Handling Pose Uncertainty}
\label{sec:uncertainty}
We use a covariance for each buffered pose $\mathbf{T}_{\mathtt{W}\mathtt{C}_t}$, expressed in $\mathtt{W}$ and conditioned on the latest pose, whose uncertainty is set to zero, so that it represents relative uncertainty and grows for older frames.
Projecting a world-frame query point $\mathbf{q}$ into camera frame $\mathtt{C}_t$ (\secref{sec:single-frame}) under this uncertainty displaces it relative to the measured surface, so the reported distance may exceed the true one by up to this displacement, and since optimistic merging keeps the largest distance across frames, obstacles are eroded below their true extent (see \figref[b]{fig:uncertainty}).
We counter this by inflating obstacles with the pose uncertainty.
Let $\mathbf{r}_t=\mathbf{q}-\mathbf{t}_t$ be the query point's lever arm from the camera center $\mathbf{t}_t$, the translation of $\mathbf{T}_{\mathtt{W}\mathtt{C}_t}$.
For a small perturbation $\boldsymbol{\xi}_t=(\delta\mathbf{t}_t,\delta\boldsymbol{\theta}_t)\in\mathbb{R}^6$ with covariance $\boldsymbol{\Sigma}_t=\operatorname{Cov}(\boldsymbol{\xi}_t)$ applied to this pose, the first-order displacement of the query point in the camera frame is
\begin{equation}
    \delta\mathbf{r}_t = \delta\mathbf{t}_t - [\mathbf{r}_t]_\times\,\delta\boldsymbol{\theta}_t = \mathbf{J}_t\boldsymbol{\xi}_t,
    \qquad \mathbf{J}_t=\begin{bmatrix}\mathbf{I}_3 & -[\mathbf{r}_t]_\times\end{bmatrix},
\end{equation}
with covariance $\mathbf{J}_t\boldsymbol{\Sigma}_t\mathbf{J}_t^\top$.
Its largest eigenvalue bounds the displacement variance in any direction, giving a $c\sigma$ margin
\begin{equation}
    \rho_t(\mathbf{q}) = c\,\sqrt{\lambda_{\max}\!\big(\mathbf{J}_t\boldsymbol{\Sigma}_t\mathbf{J}_t^\top\big)}.
    \label{eq:uncertainty-margin}
\end{equation}
For each frame, we grow the truncation radius by $\rho_t(\mathbf{q})$ before querying and subtract $\rho_t(\mathbf{q})$ from $\phi_t(\mathbf{q})$ before max-pooling.
Since $\boldsymbol{\Sigma}_t$ is relative to the latest pose, $\rho_t$ grows with frame age, so older frames contribute less free space to the max-pooling of \secref{sec:merging} and a drifted stale observation can override a newer one only while its drift stays below $\rho_t$.

\subsection{Example Application: Reactive MPC}
\label{sec:mpc}

We integrate the query into a flatness-based path-tracking and collision-avoiding \gls{mpc} \cite{mellinger_minimum_2011}: a triple-integrator model on position with jerk input, discretized at \SI{0.1}{\second} over a \SI{2}{\second} horizon, with a velocity cost along the reference tangent and a lateral deviation cost; we use a PD controller to track the result, and yaw follows the direction of motion.
We solve by \gls{sqp} with up to 20 iterations per loop, each QP with 5 interior-point iterations of HPIPM \cite{frison2020hpipm}.
Each iteration queries $\phi$ and $\nabla\phi$ at all 20 stages in one batch and enforces the linearized half-space constraint $\phi(\bar{\mathbf{q}}) + \nabla \phi(\bar{\mathbf{q}})^\top(\mathbf{q}-\bar{\mathbf{q}}) \ge r_\text{robot}$ about the previous iterate $\bar{\mathbf{q}}$, i.e., at most $400$ queries per update.
The same queries could feed corridor-based planners (\secref{sec:corridor-methods}).

\section{Experimental Evaluation}

We benchmark our method against common representations, fly it against the strongest of them in closed loop, ablate the uncertainty handling, and deploy it in real forests.
Unless stated otherwise, depth images are $640\times480$, the buffer holds the $N=5$ frames of our deployment, and all methods but FIESTA treat unobserved space as occupied.

\subsection{Setup and Baselines}
\label{sec:exp-setup}
We run all timing-critical experiments on the companion computer of our \gls{mav}, an NVIDIA Jetson Orin NX with \SI{16}{GB} of shared memory, and compare against three representations.
The first is a mapless point cloud as in \cite{florence_nanomap_2018}, a FLANN KD-tree \cite{muja_scalable_2014} per frame with the occlusion boundaries sampled at the local footprint of a depth pixel for conservative handling of \Munknownspace.
The second comprises the voxel \glspl{esdf} voxblox \cite{oleynikova_voxblox_2017} and FIESTA \cite{han_fiesta_2019} on the CPU and the \gls{gpu}-accelerated NVBlox \cite{millane_nvblox_2024}, where FIESTA treats unobserved voxels as free and the other two as occupied.
The third is a mapless mesh over the back-projected pixel grid, which represents occlusion boundaries naturally, queried with Embree \cite{wald2014embree} and with P2M \cite{zong2023p2m}, whose fast queries require an expensive acceleration structure.
The benchmark data are recorded forest sequences from a RealSense D455 with $f_x=\SI{390}{px}$ and a depth range of \SIrange{0.7}{5}{\meter}, posed by OKVIS2-X \cite{boche2025okvis2}.
By \emph{pre-processing} we mean the time required to integrate a single depth measurement into the existing representation.
Our resolution in sensor space has no single voxel equivalent, so we sweep both, the voxel grids at \SIlist{2;5;10;20}{\centi\meter} and ours at pooling levels $p\in\{1,3,7,14,27\}$, and match them only approximately, by the pooled pixel footprint in the middle of this range, \SI{2.85}{\meter}, which gives \SIlist{0.7;2.2;5.1;10.2;19.7}{\centi\meter}.

\subsection{Accuracy, Total Latency, and Memory}
\label{sec:exp-runtime}
\figref[a]{fig:res-frontier} reports the accuracy after a single integrated frame as the absolute error of $5\times10^6$ random points in the frustum against a full-resolution reference, the unpooled P2M mesh for conservative methods and an exhaustive nearest-neighbor search over the back-projected points for optimistic ones.
Fields are truncated at \SI{1}{\meter}, and pooled variants are compared against the same reference, so that their error includes the pooling.
Our conservative query differs from the mesh by a median of \SI{0.41}{\milli\meter} at full resolution and by \SI{7.8}{\milli\meter} at $p=27$.
The voxel grids degrade faster with decreasing resolution, and the error of NVBlox grows from \SI{7.9}{\milli\meter} at \SI{2}{\centi\meter} to \SI{107}{\milli\meter} at \SI{20}{\centi\meter}.
The KD-tree containing the same points as our method matches ours to within \SI{0.4}{\milli\meter} at every pooling level, as do meshes when unpooled.

\figref[b]{fig:res-frontier} reports total latency, from the start of the last of $N=5$ integrations to the answer of the $400$th query, and \figref[c]{fig:res-frontier} resolves it over the query count.
Our method answers in \SIrange{3.3}{6.2}{\milli\second} at every pooling level, with a median pre-processing of \SIrange{0.26}{0.37}{\milli\second} per frame, whereas the voxel size of the grids trades the accuracy above for latency.
NVBlox falls from \SI{199}{\milli\second} at \SI{2}{\centi\meter} to \SI{11}{\milli\second} at \SI{20}{\centi\meter} and voxblox from \SI{18.6}{\second} to \SI{146}{\milli\second} with FIESTA in between, almost all of it pre-processing.
The mapless baselines are slower than our query at every pooling level we measured, because each builds an acceleration structure per frame.
The two panels make the comparison independent of any pairing, since our finest setting is more accurate and faster than every voxel configuration, whereas our coarsest pooling level matches the finest voxel grids in accuracy at lower latency.

The advantage ends far beyond the reactive regime, since the earliest crossing at matched resolution, NVBlox at \SI{10}{\centi\meter} against $p=14$, lies at $\approx9.3\times10^3$ queries, more than an order of magnitude above the $400$ of an \gls{mpc} update, and no mapless baseline becomes cheaper up to $10^6$. On the more powerful Jetson AGX Thor, every crossing lies about three to five times higher.
\figref[d]{fig:res-frontier} reports memory, where our buffer holds \SI{5.9}{\mebi\byte} for five full-resolution frames, against \SI{66}{\mebi\byte} for NVBlox at \SI{2}{\centi\meter} and \SI{9.6}{\mebi\byte} for the KD-tree, and the gap to NVBlox widens to a factor of 44 at the \SI{10}{\centi\meter} pairing.
FIESTA is the outlier at \SI{1.4}{\gibi\byte}, because it allocates a dense volume over the mapped extent, which rules out \SI{2}{\centi\meter} on the Orin NX for this experiment and everything finer than \SI{20}{\centi\meter} in the forests of \secref{sec:exp-sim}.

\subsection{Closed-Loop Simulation in Laser-Scanned Forests}
\label{sec:exp-sim}
\begin{figure}[t]
    \centering
    \includegraphics[width=\dimexpr\linewidth-84.73bp-4pt\relax]{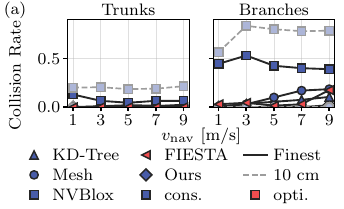}\hfill%
    \includegraphics[width=84.73bp]{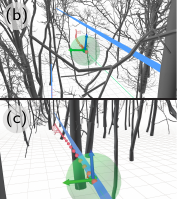}
    \caption{(a)~Collision rate over commanded speed (\simmode{onboard}) at each method's finest flyable setting (solid, NVBlox at \SI{5}{\centi\meter}) and at about \SI{10}{\centi\meter} (dashed), blue for conservative and red for optimistic methods, ours at most \SI{2}{\percent}. (b)~Branches and (c)~trunks of a laser-scanned forest.}
    \label{fig:sim-env}
\end{figure}
\begin{figure*}[t]
    \centering
    \begin{minipage}[t]{0.49\textwidth}
        \centering
        \includegraphics[width=\linewidth]{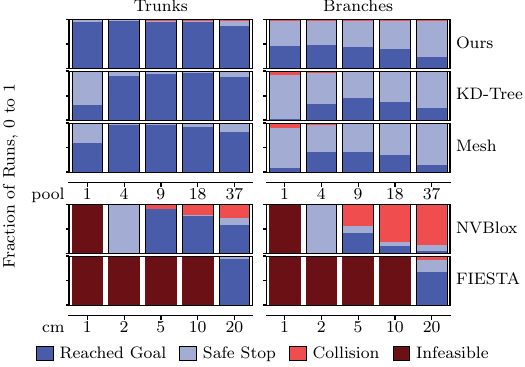}
        \captionsetup{margin=4pt}
        \caption{Outcomes in the same \gls{mpc} with every representation charged its measured cost on the Orin NX (\simmode{onboard}). Dark red marks settings the Orin NX cannot run, and each bar averages the five speeds.}
        \label{fig:res-ablation-lockstep}
    \end{minipage}\hfill%
    \begin{minipage}[t]{0.49\textwidth}
        \centering
        \includegraphics[width=\linewidth]{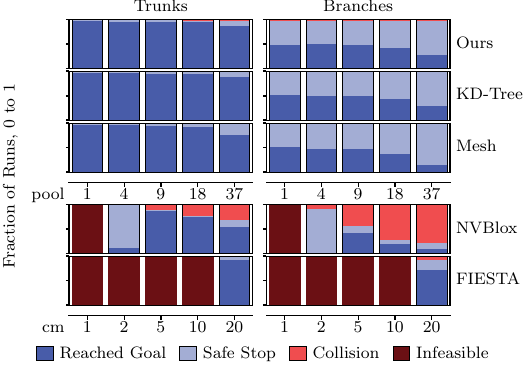}
        \captionsetup{margin=4pt}
        \caption{The sweep of \figref{fig:res-ablation-lockstep} with no compute charged (\simmode{instant}). Runs with the KD-tree, and with the mesh up to $p=9$, now match ours, so their failure was cost, while those with the voxel grids still collide.}
        \label{fig:res-ablation-zero}
    \end{minipage}%
\end{figure*}
We fly four of the benchmarked representations in closed loop with the \gls{mpc} of \secref{sec:mpc} behind the same query interface, the KD-tree and the Embree mesh over the same pooling ladder as ours, NVBlox as the fastest voxel grid, able to run as fine as \SI{2}{\centi\meter} on the Orin NX, and FIESTA at the \SI{20}{\centi\meter} its memory admits.

Each world places 50 laser-scanned trees \cite{calders_laser_2022} on a $40\times40$\,\si{\meter} plot (\figref[b and c]{fig:sim-env}), with twigs below \SI{1}{\centi\meter} in diameter removed, since a vehicle of this size cuts rather than crashes into them.
We render noise-free depth at $640\times480$ and \SI{15}{\hertz} with $f_x=\SI{525}{px}$, clamped to \SI{10}{\meter}, which leaves room to brake from \SI{9}{\meter\per\second}, and since this range puts more surface into every frame than the benchmark of \secref{sec:exp-runtime}, we charge the latencies measured here.
As before, voxel sizes and pooling levels match only approximately, and for this camera $p\in\{1,4,9,18,37\}$ spans \SIrange{0.5}{20}{\centi\meter}.
Each run tracks a straight reference through the bare trunks (\figref[c]{fig:sim-env}) or the branches in the canopy (\figref[b]{fig:sim-env}).
There is no global planner and every reference intersects geometry, so the vehicle has to find a way around it or come to a safe stop, and a perfect goal rate is not always feasible.
A run collides when a sphere of \SI{0.4}{\meter} radius about the vehicle touches geometry, stops when it moves slower than \SI{0.1}{\meter\per\second} for \SI{3}{\second}, and otherwise crosses the plot and reaches the goal.
We fly 200 runs per configuration at each of \SIlist[list-units=single]{1;3;5;7;9}{\meter\per\second} and report collision and goal rates in percent as (collision~/~goal), averaged over speeds, with the safe stops making up the remainder to \SI{100}{\percent}.

A plan computed at simulated time $t$ is released to the controller at $t+\ell$, and a representation cannot be queried while it integrates a measurement. In the \simmode{onboard} model, $\ell$ and the integration time are those the method actually incurs on the Orin NX.
The \simmode{instant} model charges no compute, which isolates the geometry of a representation from its cost, and the \simmode{delayed} model blocks the representation for a prescribed $\ell$ after every measurement.
NVBlox runs conservatively, every \gls{esdf} is truncated at \SI{0.7}{\meter}, and the \gls{mpc} keeps a clearance of $r_\text{robot}=\SI{0.5}{\meter}$.
\subsubsection{Representations in the Loop}
\figref{fig:res-ablation-lockstep} reports the outcomes in the \simmode{onboard} model.
Through the \emph{trunks} the runs with our full-resolution query reach nearly every goal without a collision (\SI{0.0}{\percent}~/~\SI{97.0}{\percent}), and no pooling level exceeds a collision rate of \SI{2.0}{\percent}.
Runs with the KD-tree (\SI{0.9}{\percent}~/~\SI{31.5}{\percent}) and with the mesh (\SI{1.1}{\percent}~/~\SI{60.3}{\percent}) at the same resolution reach fewer goals, because their \SI{169}{\milli\second} and \SI{110}{\milli\second} of pre-processing per frame delay every plan.
The vehicle then stops rather than crashes, which only coarser pooling avoids, at the cost of the resolution our method retains.
Runs with NVBlox fail at both ends of its range, since at \SI{2}{\centi\meter} one integration takes about \SI{0.8}{\second} and no run reaches the goal, and from \SIrange{5}{20}{\centi\meter} integration fits the camera period while the collision rate rises with the voxel size from \SI{5}{\centi\meter} (\SI{7.4}{\percent}~/~\SI{91.0}{\percent}) to \SI{20}{\centi\meter} (\SI{27.4}{\percent}~/~\SI{59.3}{\percent}).
FIESTA at \SI{20}{\centi\meter} (\SI{0.0}{\percent}~/~\SI{94.6}{\percent}) is the only baseline whose finest setting comes close to our goal rate.
Through the \emph{branches} most references are infeasible and a stop is the correct outcome.
Our runs stop where the vehicle cannot pass and rarely collide (\SI{0.4}{\percent}~/~\SI{47.1}{\percent}), whereas the runs with the KD-tree (\SI{5.5}{\percent}~/~\SI{4.5}{\percent}) and with the mesh (\SI{9.8}{\percent}~/~\SI{8.6}{\percent}) stop in the majority of cases and those with NVBlox collide in \SIrange{44}{81}{\percent}.
The runs with FIESTA (\SI{5.9}{\percent}~/~\SI{69.2}{\percent}) reach more goals than ours, because unobserved space is free in FIESTA and its map spans the whole scene rather than the local field of view.
FIESTA comes at fifteen times our collision rate and only a single usable resolution.

\figref[a]{fig:sim-env} resolves the \simmode{onboard} outcomes by speed at each method's finest setting.
The outcomes of the representations that delay the plan degrade with speed, since the collision rates with the KD-tree and the mesh at full resolution rise from at most \SI{1}{\percent} at \SI{1}{\meter\per\second} to \SI{11}{\percent} and \SI{19}{\percent} at \SI{9}{\meter\per\second} in the branches and with FIESTA to \SI{18}{\percent}, whereas ours stay at or below \SI{2}{\percent} at every speed, so low total latency matters most where speed leaves the least time to react.
The onboard outcomes therefore confound two causes, the geometry a representation resolves at its resolution and the total latency it incurs, which the two experiments that follow are designed to separate.

\subsubsection{Separating the Effect of Resolution}
\figref{fig:res-ablation-zero} repeats the sweep in the \simmode{instant} model, which charges no compute, so that only geometry differs.
Our outcomes in the trunks (\SI{0.0}{\percent}~/~\SI{97.7}{\percent}) and in the branches (\SI{0.4}{\percent}~/~\SI{50.1}{\percent}) move by at most three percentage points, suggesting that our cost was already negligible.
Runs with the KD-tree now match ours at every pooling level and those with the mesh up to $p=9$, so at equal resolution the mapless representations are geometrically equivalent and their divergence in \figref{fig:res-ablation-lockstep} came from cost.
Runs with the voxel grids do not recover. The collisions with NVBlox persist from \SIrange{5}{20}{\centi\meter}, where the discretized surface seems too coarse, and at \SI{2}{\centi\meter} its block budget per integration does not cover a frame, so the field stays incomplete and nearly every run stops in the trunks (\SI{0.8}{\percent}~/~\SI{12.9}{\percent}) and in the branches (\SI{7.8}{\percent}~/~\SI{0.3}{\percent}).
Runs with FIESTA (\SI{6.8}{\percent}~/~\SI{72.8}{\percent}) repeat their \simmode{onboard} outcome.
Coarser pooling has a different effect, since at $p=37$ (\SI{1.6}{\percent}~/~\SI{87.0}{\percent}) fewer runs reach the goal than at $p=1$ (\SI{0.0}{\percent}~/~\SI{97.7}{\percent}), because the inflated obstacles convert goals into stops rather than collisions.

\subsubsection{Separating the Effect of Total Latency}
\begin{figure}[t]
    \centering
    \includegraphics[width=\linewidth]{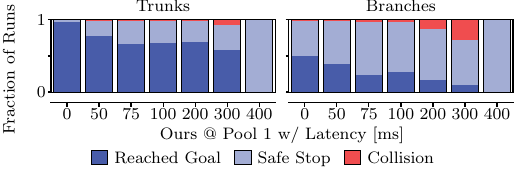}
    \caption{Our $p=1$ query blocked for $\ell$ after each measurement (\simmode{delayed}). Stops rise from \SI{50}{\milli\second} on.}
    \label{fig:latency-ablation}
\end{figure}
To charge only the cost, we fly our full-resolution query in the \simmode{delayed} model and sweep $\ell$ from \SIrange{0}{400}{\milli\second} (\figref{fig:latency-ablation}).
From \SI{50}{\milli\second} on, more runs stop through the trunks (\SI{0.1}{\percent}~/~\SI{78.4}{\percent}), the goal rate then settles near \SI{70}{\percent} up to \SI{200}{\milli\second}, collisions set in at \SI{300}{\milli\second} (\SI{7.2}{\percent}~/~\SI{58.1}{\percent}), and at \SI{400}{\milli\second} every run stops, as with NVBlox at \SI{2}{\centi\meter} onboard.
Between \SI{100}{} and \SI{200}{\milli\second}, which bracket the mesh's \SI{110}{\milli\second} and the KD-tree's \SI{169}{\milli\second}, runs still reach about \SI{70}{\percent} of the goals through the trunks with under \SI{1}{\percent} collisions, against the mesh's own onboard outcome (\SI{1.1}{\percent}~/~\SI{60.3}{\percent}) and the KD-tree's (\SI{0.9}{\percent}~/~\SI{31.5}{\percent}), so the delay explains most of the mesh's onboard loss but less than half of the KD-tree's, whose slower queries occupy the device as well.
The collisions that remain at zero delay stem from PD tracking error and the obstacle-intersecting reference, not from the planned horizons, which stay collision-free.
Taken together, runs with the voxel grids collide with or without their compute charged, so they fail through resolution, whereas runs with the mapless baselines match ours without compute charged, fail onboard through total latency, and recover only from $p\approx9$ on, while our query runs unpooled at lower latency and memory (\figref{fig:res-frontier}).

\subsection{Ablation of Uncertainty Handling}
\label{sec:exp-uncertainty}
\begin{figure}[t]
    \centering
    \includegraphics[width=\dimexpr\linewidth-84.73bp-4pt\relax]{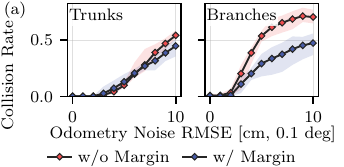}\hfill%
    \includegraphics[width=84.73bp]{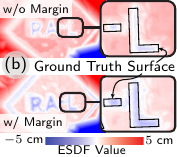}
    \caption{(a)~Collision rate on noisy D455 depth over odometry noise of $k$\,\si{\centi\meter} and $0.1k$\si{\degree} RMS, with and without the margin. The band spans the speeds. (b)~The queried \gls{esdf} around 3D-printed letters, without (top) and with margin (bottom).}
    \label{fig:uncertainty}
\end{figure}

\enlargethispage{\baselineskip}%
In the same forests we perturb the pose of each depth frame with Gaussian odometry noise of $k$\,\si{\centi\meter} and $0.1k$\si{\degree} RMS in position and orientation up to $k=10$, add the D455's depth noise \cite{keselman_intel_2017}, and fly all speeds and flight levels with and without the margin at $c=4$ (\figref[a]{fig:uncertainty}).
Below $k=2$ the two are about equally safe, because the margin is proportional to the covariance and vanishes with it.
Above that level the margin lowers the collision rate through the branches from \SI{38.2}{\percent} to \SI{21.1}{\percent} at $k=4$, whereas through the open trunks, with few fine obstacles, it pays off only above $k=6$.
Collisions still rise with $k$, because the same odometry also feeds the planner, which a margin on the observations does not repair.
Removing the depth noise shifts the pooled collision rates by under two percentage points without the margin and by at most 0.2 with it.
\figref[b]{fig:uncertainty} shows real data from a D455 posed by OKVIS2-X \cite{boche2025okvis2}, where free space enters 3D-printed letters only without the margin.

\subsection{Real-World Forest Flights}
\label{sec:exp-realworld}
\begin{figure}[t]
    \centering
    \includegraphics[width=\dimexpr\linewidth-84.73bp-4pt\relax]{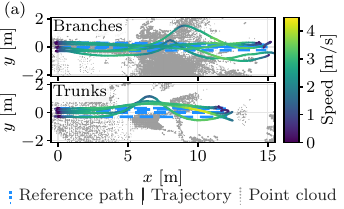}\hfill%
    \includegraphics[width=84.73bp]{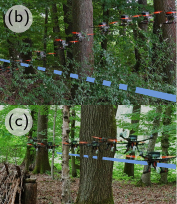}
    \caption{(a)~Five trajectories per site, colored by speed, with the reference in blue and the onboard point cloud in gray. Line and point width encode height above ground. The \gls{mav} at several instants of a run through (b)~the branches and (c)~the trunks, with an approximate reference in blue.}
    \label{fig:forest-sites}
\end{figure}
\begin{figure}[t]
    \centering
    \includegraphics[width=0.90\linewidth]{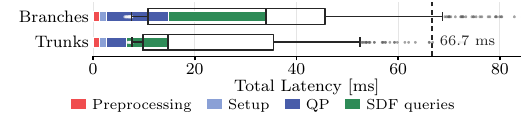}
    \caption{Onboard latency from measurement to released plan over the \num{1240} planning loops of the ten flights, with $N=5$ and $p=4$. Boxes span the quartiles, whiskers the 5th to 95th percentile, and stacked bars the median per stage.}
    \label{fig:field-latency}
\end{figure}
We deploy the full system on an \gls{mav} (\figref[b and c]{fig:forest-sites}) with a Jetson Orin NX, a RealSense D455, and OKVIS2-X \cite{boche2025okvis2}, with the depth stream clamped to the D455's usable range of \SI{5}{\meter} in these forests, min-pooled by $4\times4$ to $160\times120$, in a buffer of $N=5$ frames.
We fly five runs at each of two sites, through bare trunks (\figref[c]{fig:forest-sites}) and through bushes and fine branches (\figref[b]{fig:forest-sites}), following straight references at commanded speeds of up to \SI{4}{\meter\per\second}.
All ten flights reach the goal without a collision and without operator intervention, at a median speed of \SI{2.5}{\meter\per\second} (\figref[a]{fig:forest-sites}).
Integration and planning take a median of \SI{14.8}{\milli\second} in the trunks and \SI{34.1}{\milli\second} in the branches (\figref{fig:field-latency}), within the \SI{66.7}{\milli\second} camera period, and \SI{0.8}{\percent} and \SI{6.0}{\percent} of measurements miss it, upon which the controller tracks the previous plan.
With the full stack running, our pre-processing takes a median of \SI{1.2}{\milli\second} per frame at both sites, and the extra cost of the branches falls on the optimizer, not our query, which holds at \SIlist[list-units=single]{53;55}{\micro\second} per point, while the finer geometry needs a median of 20 \gls{sqp} linearizations against 6 in the trunks, each of which re-queries the horizon.

\section{Conclusion}
\label{sec:conclusion}

\enlargethispage{\baselineskip}%
We presented a GPU-accelerated mapless representation that answers \gls{esdf} queries in sensor space on posed depth images, with distances, gradients, conservative and optimistic handling of unknown space, and a pose-uncertainty margin, at negligible pre-processing.
On a Jetson Orin NX it answers the up to $400$ queries of a reactive planner in under \SI{7}{\milli\second} at every resolution, and at full resolution it is more accurate and faster than every voxel-grid configuration measured and one to two orders below KD-tree and mesh baselines, at the least memory at matched resolution.
In simulated laser-scanned forests at up to \SI{9}{\meter\per\second}, runs of an \gls{mpc} with our representation collide in \SI{0.4}{\percent} through the branches, against \SI{5.9}{\percent} with FIESTA and \SIrange{44}{81}{\percent} with NVBlox, and removing compute shows that the mapless baselines are geometrically equivalent to ours and fail through their total latency, which only coarser pooling avoids, while the voxel grids fail through their resolution.
Runs with FIESTA reach more goals than ours because unobserved space is free in it and its map has global extent, at fifteen times our collision rate and an allocation of \SI{1.4}{\gibi\byte}.
At two sites, all ten flights at commanded speeds of up to \SI{4}{\meter\per\second} were collision-free.
Extending the query to longer-range sensors such as LiDAR, by treating a scan as a range image, is future work.

\section*{Acknowledgment}
We thank Xinyi Chen for the support during the field trials.
The authors used generative AI tools to assist with language refinement and code generation.

\balance
\begingroup
\bibliographystyle{IEEEtran}
\bibliography{references}
\endgroup

\end{document}